\documentclass[10pt,twocolumn,letterpaper]{article}
\pdfoutput=1 
\usepackage[pagenumbers]{cvpr} 

\usepackage{graphicx}
\usepackage{amsmath}
\usepackage{amssymb}
\usepackage{booktabs}

\usepackage{paralist}
\usepackage[ruled,linesnumbered]{algorithm2e}
\usepackage{amsfonts,amssymb}
\usepackage{bbm}
\usepackage[table,xcdraw]{xcolor}
\usepackage{multirow}
\usepackage{epstopdf}
\usepackage{setspace}
\usepackage{color}

\usepackage[pagebackref,breaklinks,colorlinks]{hyperref}

\usepackage[capitalize]{cleveref}
\crefname{section}{Sec.}{Secs.}
\Crefname{section}{Section}{Sections}
\Crefname{table}{Table}{Tables}
\crefname{table}{Tab.}{Tabs.}

\def\cvprPaperID{2326} 
\def\confName{CVPR}
\def\confYear{2022}

\begin{document}

\title{Unsupervised Learning of Accurate Siamese Tracking}

\author{Qiuhong Shen\textsuperscript{1,\footnotemark[2]},
        Lei Qiao\textsuperscript{2},
        Jinyang Guo \textsuperscript{3},
        Peixia Li \textsuperscript{3},
        Xin Li \textsuperscript{4}, \\
        Bo Li \textsuperscript{2},
        Weitao Feng  \textsuperscript{3},
        Weihao Gan \textsuperscript{2,5},
        Wei Wu \textsuperscript{2,5},
        Wanli Ouyang \textsuperscript{3,5} \\
        \textsuperscript{1} Harbin Institute of Technology (Shenzhen) 
        \textsuperscript{2} SenseTime Research \\
        \textsuperscript{3} The University of Sydney 
        \textsuperscript{4} Peng Cheng Laboratory 
        \textsuperscript{5} Shanghai AI Laboratory \\
        \tt\small{
            \{shenqiuhong0905,xinlihitsz\}@gmail.com,
            \{qiaolei,libo,ganweihao,wuwei\}@sensetime.com,}
        \\
        \tt\small{\{jinyang.guo,peixia.li,weitao.feng,wanli.ouyang\}@sydney.edu.au
        }
}


\maketitle

\renewcommand{\thefootnote}{\fnsymbol{footnote}}
\footnotetext[2]{Work performed when Qiuhong Shen is an intern at SenseTime.}

\begin{abstract}
Unsupervised learning has been popular in various computer vision tasks, including visual object tracking. However, prior unsupervised tracking approaches rely heavily on spatial supervision from template-search pairs and are still unable to track objects with strong variation over a long time span. As unlimited self-supervision signals can be obtained by tracking a video along a cycle in time, we investigate evolving a Siamese tracker by tracking videos forward-backward. 
We present a novel unsupervised tracking framework, in which we can learn temporal correspondence both on the classification branch and regression branch. Specifically, to propagate reliable template feature in the forward propagation process so that the tracker can be trained in the cycle, we first propose a consistency propagation transformation. We then identify an ill-posed penalty problem in conventional cycle training in backward propagation process. Thus, a differentiable region mask is proposed to select features as well as to implicitly penalize tracking errors on intermediate frames. Moreover, since noisy labels may degrade training, we propose a mask-guided loss reweighting strategy to assign dynamic weights based on the quality of pseudo labels. In extensive experiments, our tracker outperforms preceding unsupervised methods by a substantial margin, performing on par with supervised methods on large-scale datasets such as TrackingNet and LaSOT. Code 
is available at \href{https://github.com/FlorinShum/ULAST}{https://github.com/FlorinShum/ULAST}.
\end{abstract}

\section{Introduction}
Visual tracking has become an integral part of various video applications such as autonomous driving and video recognition.
In the mainstream of visual object tracking, deep learning-based trackers are dominant~\cite{tracking_survey21}, requiring a large number of labeled videos. 
Since the labeled data occupy a relatively small portion of practical scenes, the trained tracker cannot reliably track previously unseen objects. Thus, learning from unlabeled videos becomes a promising approach.
Prior works on unsupervised tracking fall into two categories: exploiting self-supervision signals in videos from either the spatial or temporal dimensions. 
For the first category~\cite{s2siamfc, croptransformpaste}, the focus is on how to construct a template-search pair using a still frame.
Since these methods are limited by the inability to learn temporal correspondence over long periods of time, trained trackers can no longer track objects with strong variation.
To cope with the appearance variations that occur during online tracking, we center our attention on methods in the latter category, exploiting the temporal self-supervision signal in videos.

In supervised tracking methods, box-regression branch has been demonstrated to be effective to capture the objects with large scale variation along temporal dimension~\cite{SiamRPN, ATOM}. 
However, in existing unsupervised methods, this branch is always absent~\cite{s2siamfc, udt, pul}. Recently, USOT~\cite{usot} introduced a box-regression head, but the tracker is initially trained with template-search pairs from single frames, followed by cycle memory training to enhance the robustness of classification branch, whereas the box-regression branch is trained with spatial supervision alone.
In this paper, we aim to train a better tracker by learning temporal correspondence both on the classification branch and on the regression branch.
However, we identify there are three critical challenges.
\begin{figure}[t]
\centering
\includegraphics[width=0.8\linewidth]{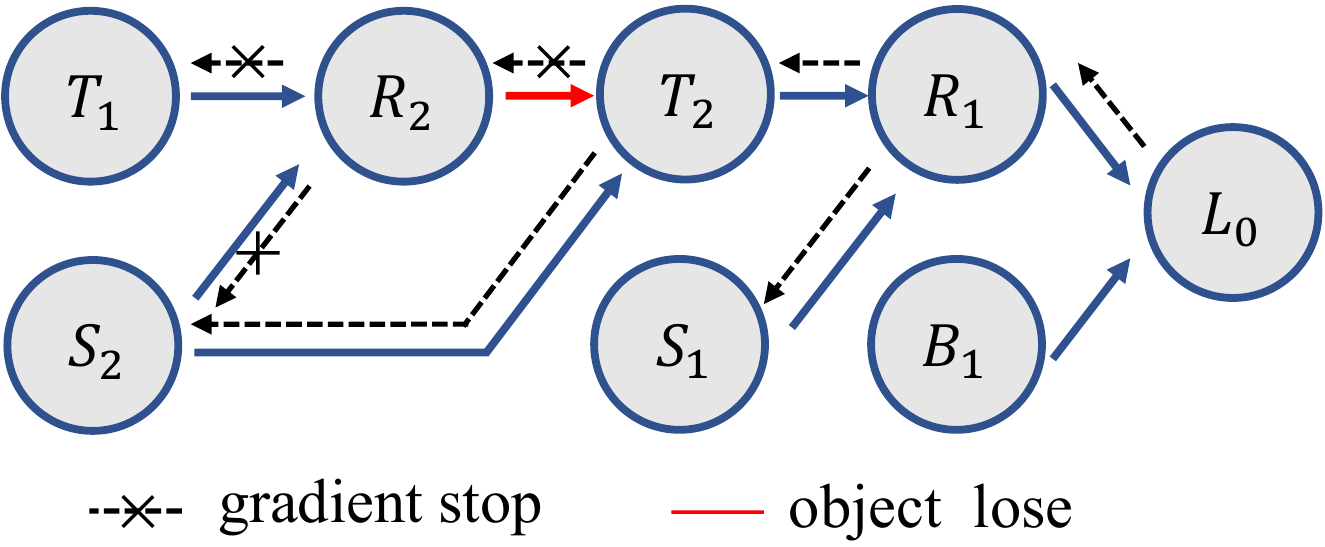}
\vspace{-3mm}
\caption{Three challenges when learning a better tracker with temporal correspondence on both classification and regression branches. First, in forward propagation process, tracker often loses target objects in intermediate frames, breaking down the training pipeline. Second, in backward propagation process, gradient cannot flow through the whole framework due to the RoI-Align. Third, the pseudo labels are often noisy, degrading the performance. In this figure, $T$, $S$, $R$, $B$, and $L$ denote template kernel, search region, candidate boxes, pseudo labels, and loss value, respectively. The subscript indexes the temporal order of frames.
}
\label{fig:gd_stop}
\vspace{-5mm}
\end{figure}

First, despite unlimited self-supervision can be obtained by tracking a video along a cycle in time, how to explore the self-supervision signal in temporal dimension of videos for training a tracker equipped with a box-estimation branch is not well explored in existing methods.
In cycle training, as illustrated in Fig.~\ref{fig:gd_stop}, a tracker is assumed to be capable of tracking back to the initial location. The tracker is evolved in the cycle by utilizing the inconsistency between start and end location in initial frames.
However, when training from scratch, it is hard for the tracker to find the target object. 
The template kernel generated on the intermediate frames is likely to not contain any features of the target object, which means the tracker cannot return to the initial location. The training pipeline will break down after several iterations.
Second, we identify misalignment in cycle training: 
when the tracking result $R_2$ is inaccurate, the generated template kernel $T_2$ is likely to embrace many distractor features, but the imposed loss still forces the tracker to predict accurate target box by using such noisy template, which causes an ill-posed penalty.
From the gradient flow of cycle training, we can observe the select operation like RoI-Align, since the coordinate is quantized, is not differentiable on boxes coordinate. Therefore the gradient cannot back propagate to the node before $R_2$.
In other words, the tracking errors on intermediate frames cannot be penalized in this pipeline.

Third, 
as unsupervised tracking framework still relies on the pseudo labels in initial frames
%
and box-regression branch training requires objects that have clear edges, pseudo labels in initial frames are crucial. However, we observe these labels are likely to be noisy, degrading the tracking performance. 

To address the aforementioned three challenges, we propose a novel unsupervised tracking framework called ULAST, which aims to learn temporal correspondence both on the classification branch and regression branch. It consists of three newly proposed components: consistency propagation transformation, region mask operation, and mask-guided loss re-weighting. Specifically, the consistency propagation transformation aims to generate reliable template kernel for tracking next frame, which uses both long-term and short-term information from template kernels and search regions of previous frames. As a result, it enables our framework to exploit temporal self-supervision signal and avoiding the training pipeline breaking down. Instead of using RoI-Align, our region mask operation selects features with all candidates in $R_2$ based on the search region feature and predicted bounding boxes from previous frame, and makes regression and classification heads differentiable to implicitly penalize tracking errors on intermediate frames. The mask-guided loss re-weighting strategy dynamically assigns weights to samples based on the quality of their pseudo labels, which avoid using the noisy pseudo labels.

We evaluate the trained tracker on five diverse benchmark datasets, and the favourable performance against state-of-the-art methods demonstrate the effectiveness of our proposed framework. The main contribution of this work are summarized as follows:
\begin{itemize}
\setlength{\itemsep}{0pt}
\setlength{\parsep}{0pt}
\setlength{\parskip}{0pt}

%
\item We propose a novel unsupervised learning framework called ULAST, which can lean temporal correspondence both on classification and regression branches.

\item A consistency propagation transformation is proposed to generate reliable template kernel, avoiding the training process of our ULAST framework breaking down.

\item A differentiable region mask operation is proposed to select features as well as implicitly penalize the tracking errors of intermediate frames in backward propagation process.

\item A mask-guided loss re-weighting strategy is proposed to mitigate the negative impact of noise on training.
\end{itemize}

\section{Related work}
\noindent{\textbf{Supervised visual tracking}}
The past few years have witnessed significant performance improvement in deep-learning based trackers. 
We can roughly divide these trackers into two categories: online-optimized trackers~\cite{eco, ATOM, DiMP, PrDiMP} and offline trackers~\cite{SiamRPN++, siamatt,SiamDW, SiamCAR,TransT, SiamGAT, stark}.
Online-optimized trackers rely on online update methods with dedicated design. The rough position of target objects are figured out by ridge regression with updated template kernels. Then refinement is applied to estimate accurate bounding boxes.
On the other hand, offline trackers learn to match the template and search region in metric space instead.
This category is dominated by Siamese-network based trackers.
The pioneering work SiamFC~\cite{SiamFC} extracts features of template and search patches with shared backbone network. Then cross correlation is applied to generate response map for locating the target.
Various efforts have been made in this category, such as better backbone networks~\cite{SiamRPN++}, target-aware attentions~\cite{siamatt, TADT}, anchor-free regression~\cite{SiamDW, SiamCAR}, and effective template-search fusion~\cite{l2fuse, l2match}.
Efficient tracking~\cite{lightrack, chase} with Siamese-based trackers is also explored with pruning~\cite{ChannelPrune, MultiPrune} and network architecture search~\cite{EffSearch}.
Nevertheless, all of these methods require extensive supervised training involving a large number of annotated videos to learn the correspondence between template and search region.
In contrast, our work is an unsupervised learning framework for accurate Siamese-based tracking, which does not require a huge annotation cost.

\begin{figure*}[tbp]
\centering
\includegraphics[width=1.0\linewidth]{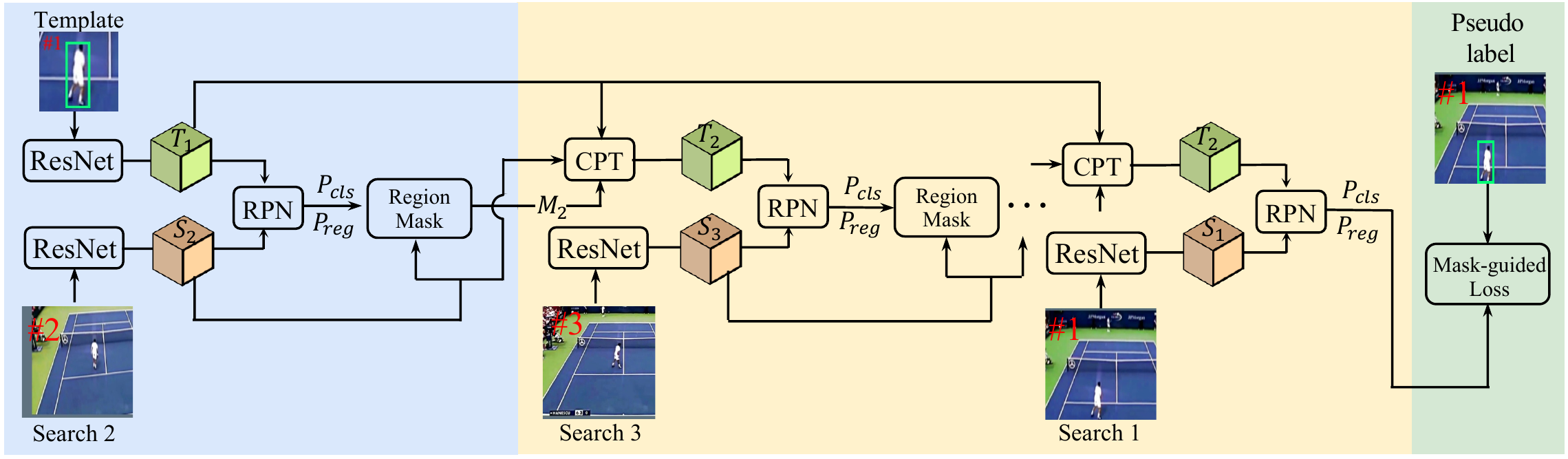}
\vspace{-8mm}
\caption{\textbf{Overview of our Framework}. 
Here we illustrates the overall framework with tracking 3 frames in $1 \rightarrow 2 \rightarrow 3 \rightarrow 2 \rightarrow 1$ order.
}
\label{fig:net_arch}
\vspace{-6mm}
\end{figure*}

\noindent{\textbf{Unsupervised visual tracking. }}
Since it is costly to collect annotations for videos, unsupervised tracking~\cite{ludt, s2siamfc, pul, usot} becomes a promising approach for training more robust trackers. 
The pioneering work UDT~\cite{udt} trained a Discriminative Correlation Filters (DCF) based tracker by forward-backward tracking frames with the supervision of consistency loss.
These works indicated that cycle consistency in videos can be utilized to effectively train a robust tracker by forward-backward tracking multiple frames.
On the other hand, $s^2$siamfc~\cite{s2siamfc} proposed a Siamese network based unsupervised training framework by mining the self-supervision in single frames, in which adversarial masking is learnt to construct template-search pairs from identity frame.
However, tracking performance of these methods~\cite{usot, udt} depends heavily on online updating schemes.
Without online updates, these unsupervised trained trackers cannot handle challenging objects with dramatic variation.
Recently, acquiring self-supervision signals from both spatial and temporal dimension becomes the promising way for unsupervised tracking. 
Zheng \textit{et al.} ~\cite{usot} proposed an unsupervised training approach by naive training from single frame in first stage, then cycle training is adopted to learn on longer temporal spans.
PUL~\cite{pul} proposed to initially learn a background discrimination model by contrastive learning. Then the model continues training with temporal corresponding patches mined with a noise-robust loss.
Besides, various pretext tasks~\cite{timecycle,randomwalk,videocp} are constructed in learning visual representation from videos by utilizing the forward-backward tracking idea.
Different from these works, our framework mainly focuses on learning classification and regression capability simultaneously from the temporal supervision, achieving superior tracking performance.

\section{Proposed Method}
\subsection{Preliminary}
\noindent\textbf{Siamese-network based Tracker.}
To handle scale variation of the target in the cycle, our ULAST is built upon Siamese-network based region proposal network~\cite{SiamRPN++}. 
Suppose we need to use a template patch to find the target object in a search region. The tracker first use a ResNet50 with shared parameter to extract the feature from the template and search region, and we denote the extracted feature from the template and search region patches as $T$ and $S$, respectively. Then, a region proposal network is learnt to generate the bounding box and the corresponding class. The network is optimized with classification loss and regression loss:
\begin{eqnarray}
L^{l} = \lambda_{1} L_{cls}^{l} + \lambda_{2} L_{reg}^{l},
\label{eq:base_loss}
\end{eqnarray}
where $L_{cls}^{l}$ and $L_{reg}^{l}$ are respectively the Focal loss~\cite{FocalLoss} and the $L_1$ loss. 
$\lambda_{1}$ and $\lambda_{2}$ are coefficient to balance two terms, the superscript of $L^{l}$ denotes the conventional (legacy) loss in Siamese pair training paradigm~\cite{SiamRPN}.

\noindent\textbf{Cycle Training.}
Suppose we track three sampled frames in a cycle with the given pseudo-label of the first frame.
when tracking the 2nd frame, we use the template kernel extracted from patch in the 1st frame to predict candidate boxes on search region. Then these boxes are leveraged to generate new template kernel by selecting feature from search region.
When tracking subsequent frames, we use template kernels generated from last search region to track.
Generally, we track frames in palindrome order (i.e frames are ordered as 1,2,3,2,1) to track the objects back to the first frame. The tracker can be optimized by inconsistency the final tracking result and the pseudo label in the first frame.
Formally, we call this training pipeline cycle training, as opposed to legacy training from single frames. The formulation of loss $L^{c}$ in cycle training is same as that in legacy training, which consists of Focal loss \cite{FocalLoss} and $L_{1}$ loss. Thus, the total loss in cycle training is formulated as:
\begin{eqnarray}
L_{total} = (1 - \lambda_{c} ) L^{l} + \lambda_{c} L^{c},
\end{eqnarray}

Here $L^{l}$ is the loss for self-tracking, constructing template-search pairs from first frames. The cycle loss $L^c$ is computed by tracking back to the first frame in the cycle,  $\lambda_c$ is a manual weight parameter.

\subsection{Overview}
The architecture of our framework is shown in Fig.~\ref{fig:net_arch}, in which we use with three sparse sampled video frames for illustration. 
Given a palindromic video sequence, we first use a shared ResNet50 as the backbone to extract features from template of the first frame and search patches from the second frame and generate the template feature $T_1$ and search region feature $S_2$. Then, the extracted feature $T_1$ and $S_2$ are fed into the regression branch and classification branch of the region proposal network (RPN) to yields box-regression result $P_{reg}$ and corresponding classification confidence score $P_{cls}$. Then, we pass $P_{cls}$ and $P_{reg}$ to the region mask operation and generate the regional mask $M_2$. 
After that, our CPT module takes the search feature $S_2$ and corresponding regional mask $M_2$ as input to generate the template feature for the second frame $T_2$.
At the same time, search region feature $S_3$ is generated from 3rd frame.
This search feature and the template feature $T_2$ are then fed into RPN to predict 
box classification and regression results on 3rd frame.
This process is repeated and finally generate the tracking result at the first frame, which will be then used to calculate the mask-guided loss based on the pseudo-label at the first frame.
\begin{figure}[b]
\vspace{-5mm}
\centering
\includegraphics[width=1\linewidth]{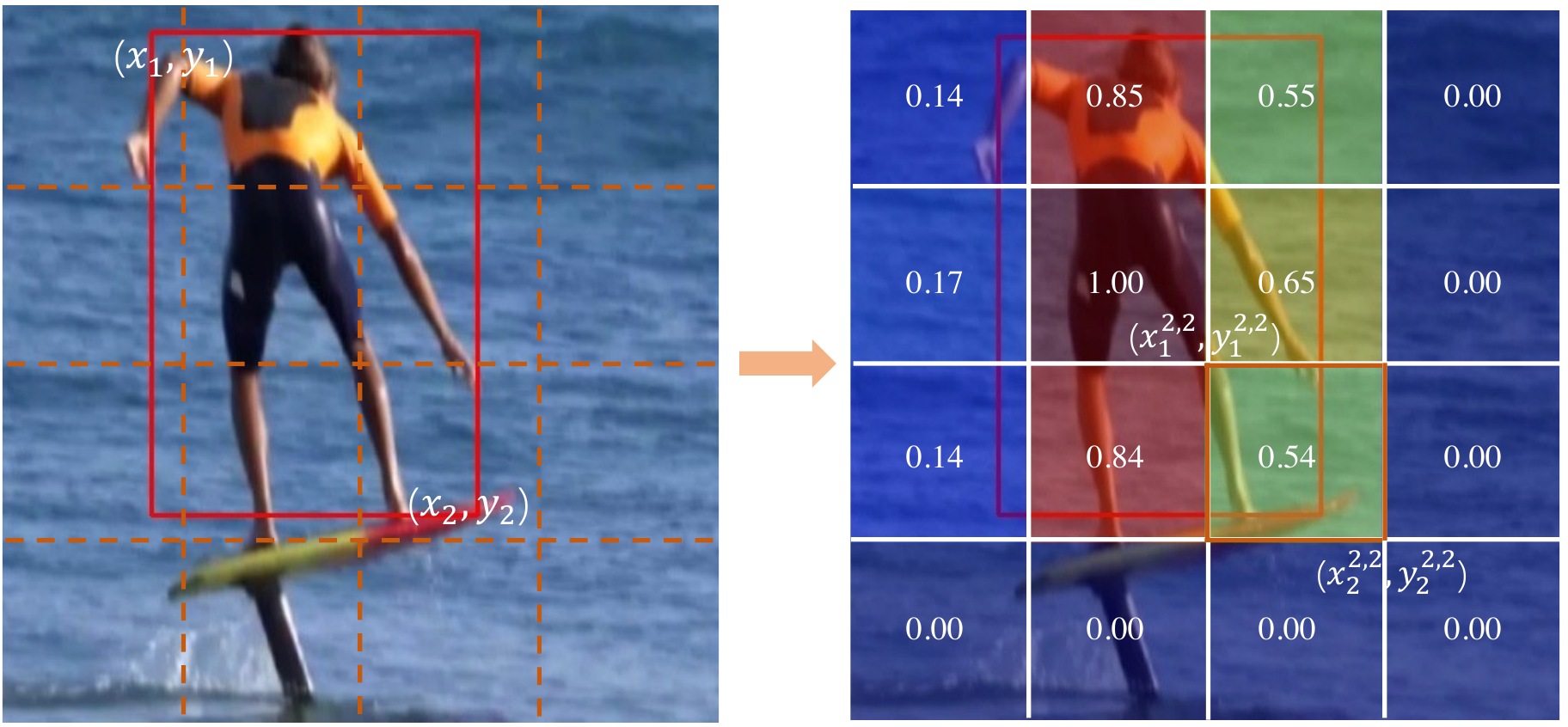}
\caption{\textbf{Region mask operation}. 
We illustrate a grid map computed for one predicted box on search region. For better presentation, the grid size is set as $4 \times 4$. Better viewed with zoom in.
}
\label{fig:region_mask}
\end{figure}

\subsection{Training with cycle-consistency}
\noindent{\textbf{Region mask.}}
\label{subsec:region_mask}
As introduced before, conventional cycle training suffered from the ill-posed penalty.
As the top-n prediction results of classification and regression branch are likely to be inaccurate in initial training stage. Generated new templates feature based on these top-n boxes may do not contain any target object feature. Forcing such template to track back to the initial target position leads to the ill-posed penalty.
And the feature selection operation like RoI-Align is not differentiable on boxes coordinates, causing intermediate frames tracking error cannot be penalized. 
Although Precise RoI-Pooling~\cite{prpool} can be used to make the coordinate differentiable, it cannot pass the gradient to all the candidate boxes. 
To address these issues, it is necessary to introduce a module to that can select target features from last search region involving all estimated boxes and be differentiable on the coordinates.

Therefore, we propose a novel operation called region mask to select region-wise features and make the classification and regression branches differentiable in cycle training.
Suppose we need to calculate the region mask based on the search region feature $S_{t} \in \mathbb{R}^{C \times H \times W}$, here subscript $t$ denotes $t \,\, th$ frame in sampled frames.
The input of the region mask operation is the output of the RPN, which consists of the output $P_{cls}$ and $P_{reg}$ from the classification and regression branch, respectively. 
We first introduce a grid of size $H \times W$, same as the spatial resolution of search region feature $S_{t}$. 
In Fig~\ref{fig:region_mask}, we illustrate the region mask operation when using a grid of $4 \times 4$ size.
Let us suppose there are totally $K$ estimated box in the output of the regression branch $P_{reg}$ and the grid map for the $k \, th$ predicted box is denoted as $G_k$. 
Denotes the $k \, th$ predicted box as $(x_1, y_1, x_2, y_2)$ and the grid box at the position $(i, j)$ as $(x_1^{ij}, y_1^{ij}, x_2^{ij}, y_2^{ij})$. The grid value $G_k^{(i, j)}$ for the grid map $G_k$ at $i \, th$ row and $j \, th$ column can be calculated as follows:
\begin{eqnarray}
\vspace{-6mm}
G^{(i, j)}_k = \frac{(\hat{x}_2^{ij} -\hat{x}_1^{ij})(\hat{y}_2^{ij} - \hat{y}_1^{ij})}{(x_2^{ij} - x_1^{ij})(y_2^{ij} - y_1^{ij})},
\vspace{-6mm}
\label{eq:rm_1}
\end{eqnarray}
where $(\hat{x}_{1}^{ij}, \hat{y}_{1}^{ij})$ and $(\hat{x}_{2}^{ij}, \hat{y}_{2}^{ij})$ are the top-left and bottom-right intersection corner, $\hat{x}_{1}^{ij} = max(x_1, x_1^{ij})$, $\hat{y}_{1}^{ij} = max(y_1, y_1^{ij})$,  $\hat{x}_{2}^{ij} = min(x_2, x_2^{ij})$ and $\hat{y}_{2}^{ij} = min(y_2, x_2^{ij})$.
Here, the grid value represents the overlap ratio of the fixed grid with predicted box.
Obviously, this formulation is naturally differentiable on the coordinates like IoU (Intersection of Union).
Therefore, the grid map for all the boxes in this search region can be collected into a set $\{G_k\}$, for $k=1,2,\dots,K$. 
As each bounding box predicted by the RPN has a corresponding confidence score, it is intuitive to combine these grid maps with their confidence.
After getting the grid map set $\{G_k \}$, we aggregate it into a single channel regional mask. 
For the same spatial grid with duplicated positive grid values, we only use the grid value from the $k \, th$ predicted box with the highest confidence score, and set the grid value from the other predicted box as 0.
With this processing, we generate a new grid map $\{ \tilde{G}_k \}$, at most only one grid map $\tilde{G}_k$ has grid value greater than 0 across $K$ grid maps for a certain spatial pixel $(i, j)$.
For final aggregation, the regional mask of the search region feature $S_{t}$ can be calculated as follows:
\begin{eqnarray}
M_{t} = \sum_{k=1}^{K}\mathbbm{1}(s_k, TH) \cdot {s_k  \tilde{G}_k },
\label{eq:rm_2}
\end{eqnarray}
where the indicator function is defined as:
\begin{equation}
\mathbbm{1}(s_k, TH) \left\{\begin{array}{ll}
1, & if \,\, s_k \geq TH \\
0, & otherwise
\end{array}\right..
\label{eq:rm_3}
\end{equation}
Here, $M_{t} \in \mathbb{R}^{H\times W}$ is the generated regional mask. $TH$ is a manually set threshold. $s_k$ denotes the classification confidence score for $k \, th$ grid map.

\begin{figure}[tbp]
\centering
\includegraphics[width=0.85\linewidth]{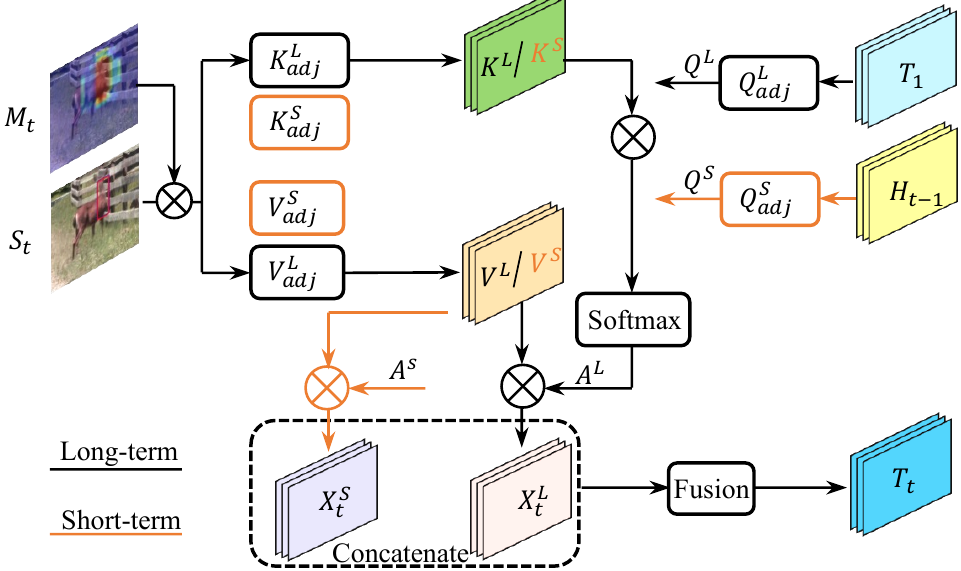}
\caption{\textbf{Consistency propagation transformation}. In this transformation, we generate reliable template kernel from the last search region by retrieving long-shot term features. Black lines denote long-term target feature retrieval while orange lines denote short-term target feature retrieval. }
\label{fig:lst_memory}
\vspace{-4mm}
\end{figure}

\noindent{\textbf{Consistency propagation transformation (CPT).}}
As the aforementioned pipeline broken problem, it is crucial to ensure consistency propagation between frames, i.e., the tracker should not lose the target object.
In our framework, we argue that the reason for this problem is the lack of exploration of temporal consistency when introducing the box-estimation branch.
The predicted boxes of subsequent frames will be largely affected by those from current frames. Namely, once the predicted box in the current frame is inaccurate, it is difficult to generate accurate predicted boxes in the subsequent frames. Therefore, we propose a consistency propagation transformation (CPT) module to exploit the self-supervision signal in the temporal dimension. Specifically, as illustrated in Fig.~\ref{fig:lst_memory}, our CPT module aims to retrieve a new template kernel $T_{t}$ from search region feature $S_{t}$ with regional mask $M_{t}$ based on the predicted result on $t \,\,th$ search frame.
Specifically, to squeeze the noisy features in the search region feature denoted as $S_{t}$, a long-short term consistency module is introduced to retrieve features from that. As the initial template feature denoted as $T_{1}$ comprises the most reliable features of the target, we get the long-term feature conditioned on $T_1 \in \mathbb{R}^{C \times h \times w}$.
First, the region mask $M_{t} \in \mathbb{R}^{H \times W}$ is multiplied with the search feature $S_{t} \in \mathbb{R}^{C \times H \times W}$ to generate a pre-selected search feature denoted as $\Tilde{S}_{t}$:
\begin{eqnarray}
\vspace{-4mm}
\Tilde{S}_{t} = S_{t} \otimes M_{t},
\vspace{-6mm}
\end{eqnarray}
where $\otimes$ denotes the element-wise multiplication. Then an an $1 \times 1 $ convolution denoted as $K_{adj}^L$ is applied to $\Tilde{S}_{t}$ for generating the long-term key features $K^L \in \mathbb{R}^{C \times H \times W}$. 
We also use another convolution denoted as $Q_{adj}^{L}$ to apply on the template feature $T_1$ and generate the long-term query feature $Q^L \in \mathbb{R}^{C \times h \times w}$. The superscript $L$ denotes the long-term. Then, we reshape the query and key into the shape of $C\times N_z$ and  $C\times N_x$ respectively, where $N_z = h\times w$ and $N_x = H \times W$. Therefore,
%
We can generate a long-term affinity matrix $A^L \in \mathbb{R}^{N_z \times N_x}$ as:
\begin{eqnarray}
A^L = Softmax_{col}((Q^L)^T K^L) \in \mathbb{R}^{N_z \times N_x},
\end{eqnarray}
where $Softmax_{col}$ is the softmax operation along the column dimension.
We also use another $ 1\times 1$ convolution denoted as $V^L_{adj}$ to adjust $\tilde{S}_{t}$ for generating the long-term value features $V^L \in \mathbb{R}^{C \times N_x}$, which will be multiplied with the affinity matrix to generate the long-term feature $X^L_{t}$:
\begin{eqnarray}
X^{L}_{t} = reshape (A^{L} (V^L)^{T}) \in \mathbb{R}^{C \times h \times w}.
\end{eqnarray}

For the short-term feature generation, the only difference from that for long-term is the input of the query adjust operation is different. We use a hidden template $H_{t-1}\in \mathbb{R}^{C \times h \times w}$ as the input of the query adjust operation and generate the short-term query feature, where $H_{t-1}$ is calculated when tracking the last frame:
\begin{eqnarray}
H_{t-1} = f_{\phi} (concat(X^{L}_{t-1}, T_1)),
\end{eqnarray}
where $concat(\cdot)$ is the concatenation operation. $f_{\phi}$ denotes the hidden template aggregation module, which is a conv-bn block with learnable parameter $\phi$.

Except for the difference mentioned above, the other operations are the same as those for long-term feature generation. We use the similar manner to generate the short-term feature $X^S_{t}$.
Finally, with the long-term and short-term feature retrieved from current search feature $S_t$, we generate the output of the CPT module, which can be formulated as:
\begin{eqnarray}
T_{t} = h_{\theta} (concat(X^{S}_{t}, X^{L}_{t})).
\end{eqnarray}
Here, $h_{\theta}$ denotes the aggregation operation implemented by using a conv-bn block with the learnable parameter $\theta$.

\begin{figure}[b]
\vspace{-4mm}
\centering
\includegraphics[width=0.9\linewidth]{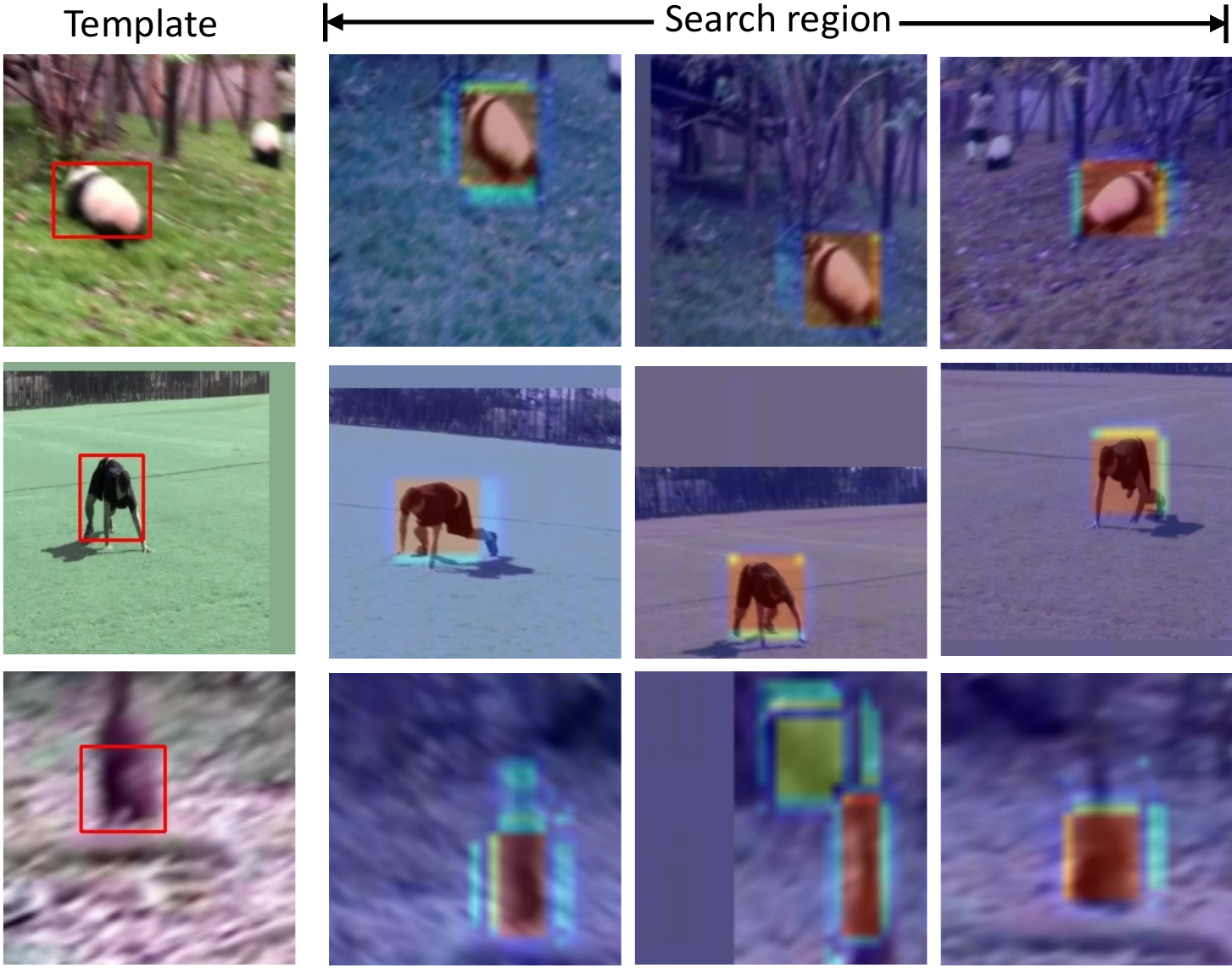}
\caption{\textbf{Visualization of regional mask.} The pseudo label sample in the 3rd row encompasses only part of the target object, which is noisy for training the tracker, resulting in regional mask with larger high response area (green part in the 3rd row frames).
}
\vspace{-4mm}
\label{fig:mask_pro}
\end{figure}

\subsection{Learn from noise label}
In our ULAST, we use the unsupervised trained optical flow model~\cite{arflow} to generate the pseudo labels of the first frame, which will be used in the training process. However, these pseudo labels often are noisy, which hinders the performance of the trained tracker.
Existing approaches~\cite{s2siamfc, pul} only use the classification confidence from the response map to filter out label noise. We argue that the classification result is not sufficient to evaluate the importance of each sample. Therefore, we also take the regression result into consideration. 
Intuitively, template kernels generated from pseudo labels with more noise often produce larger response area from background in search regions, which can be identified from Fig~\ref{fig:mask_pro}. 
To this end, we propose a mask-guided loss re-weighting strategy to re-weight the loss from the pseudo-labels of various qualities.
Specifically, suppose we have a video batch of size $B$. For the $b$-th sample in this batch, we first use the tracker to generate the classification output $P_{cls}$ and regression output $P_{reg}$ by constructing a template-search pair from the first frame. Then, based on $P_{cls}$ and $P_{reg}$, we get the regional mask $\hat{M}^b$ by using Eq. \ref{eq:rm_1}, \ref{eq:rm_2} and \ref{eq:rm_3} for this video. 
On the other hand, given the pseudo-labelled first frame, we also calculate $\bar{M}^{b}$, the only difference is that we set $K=1$ and $s_k=1$ in Eq. \ref{eq:rm_2}, as there is only one ground-truth box in the first frame in this case.
Then we learn a dynamic weight $w_b$ for each sample in a batch according to the relatively high response region in such formulation:
\begin{eqnarray}
w_b =\log_{\gamma}(\alpha - \frac{\sum_{p}{\sum_{q}\mathbbm{1}(\hat{M}^b_{p, q}, \beta)}}{\sum_{p}\sum_{q}\mathbbm{1}(\bar{M}^{b}_{p,q}, \beta)})),
\end{eqnarray}
where $\mathbbm{1}$ is an indicator function defined as:
\begin{equation}
\mathbbm{1}(\hat{M}^b_{p, q},\beta)\left\{\begin{array}{ll}
1, & if \,\, \hat{M}^b_{p, q} \geq \beta \\
0, & otherwise
\end{array}\right..
\end{equation}
Here, $\hat{M}^b_{p,q}$ and $\bar{M}^b_{p,q}$ are the grid at the $p$-th row and $q$-th column of the regional mask for the $b$-th sample in this batch. $\gamma$ denotes the factor to scale the value of $w_b$, $\alpha$ denotes threshold for assigning weights to noisy labels, and $\beta$ denotes threshold to filter out distractors of low response.
Therefore, the loss function can be formulated as:
\begin{eqnarray}
L = \frac{1}{B} \sum_{b} w_b (\lambda_{1}L_{cls} + \lambda_{2} L_{reg}),
\end{eqnarray}
where $L_{cls}$ and $L_{reg}$ are the classification loss and regression loss introduced in Eq. \ref{eq:base_loss}. Here we omit superscript of $L$ as this formulation is both applied on $L^{l}$ and $L^{c}$.

\subsection{Online tracking}
As legacy training and cycle training are both adopted in offline training phase, the input of template kernel in RPN is compatible with template feature extracted from patches or feature retrieved from historical search feature by CPT module.
For real-time tracking, our tracker can perform like SiamRPN~\cite{SiamRPN++}, operating at a high speed about 80 FPS.
For robust tracking, cycle training enables the tracker to update with a memory queue. 
Detailly, a memory queue of length $N_L$ stored with historical search features and region masks is maintained, including features of initial frame and $N_L - 1$ historical samples of the highest score. Once the memory queue is updated, the memory kernel will be updated. Besides, the hidden template $H_t$ is updated every $N_s$ frames with the highest score in this short interval.
For the trade-off between speed and accuracy, the classification map $R_{cls}$ from the legacy kernel and memory kernel is combined with $R_{cls} = (1 - \lambda_m) R_{cls}^{L} + \lambda_m R_{cls}^{M}$, $\lambda_m$ denotes the weight to balance the classification score, while the regression map is generated with the template kernel extracted from the initial template patch. Here, $R_{cls}^{L}$ denotes the response map in classification branch that applying the template kernel extracted from initial frame, while $R_{cls}^{M}$ is the classification response map when applying the template kernel extracted from online memory queue.
Specifically, We set $N_L =6$ and $N_s=10$ throughout all online updating experiments.

\section{Experiments and Results}
\subsection{Implementation details}
\noindent{\textbf{Data preparation.}}
The labels of our training are generated by using the off-the-shelf unsupervised optical flow model~\cite{arflow} on datasets Got10k~\cite{got10k}, LaSOT~\cite{LaSOT}, VID~\cite{vid} and YoutubeVOS~\cite{ytvos}, the data sampling strategy is similar to USOT~\cite{usot}. One reliable template frame and three search region frames with large temporal gaps are sampled from a video for cycle training.
The template patch is cropped as size $127 \times 127$, and the search frames are replicated as palindromes for tracking back to the initial frame. The spatial size of all input frames is resized as $640 \times 480$ without crop. 
Specifically, our framework only requires objects' initial box and subsequent center position, reducing the reliance on pseudo-labels.
As the optical flow method struggles with scale variation of target objects, in the cycle training, the search frames are cropped as $255 \times 255$ according to the input object center and last frame estimated object scale instead of the scale initialized in pseudo-labels.

\begin{table*}[t]
\setlength{\abovecaptionskip}{0cm}
\setlength{\belowcaptionskip}{0cm}
\centering
\caption{Evaluation results on TrackingNet, VOT2016 and VOT2018 benchmark datasets. Unsup denotes unsupervised training.}
\label{tab:all_performance}
\small
\begin{tabular}{cc|ccccccccc}
\hline
 &
  \multicolumn{1}{l|}{} &
  \multicolumn{3}{c}{TrackingNet} &
  \multicolumn{3}{c}{VOT2016} &
  \multicolumn{3}{c}{VOT2018} \\
\multirow{-2}{*}{Tracker} &
  \multicolumn{1}{l|}{\multirow{-2}{*}{Unsup}} &
  Suc $\uparrow$ &
  Pre $\uparrow$ &
  NPre $\uparrow$ &
  EAO $\uparrow$ &
  Acc $\uparrow$ &
  Rob $\downarrow$ &
  EAO$\uparrow$ &
  ACC$\uparrow$ &
  Rob$\downarrow$ \\ \hline
SiamFC~\cite{SiamFC} &
  No &
  0.571 &
  0.533 &
  0.663 &
  0.235 &
  0.532 &
  0.461 &
  0.188 &
  0.503 &
  0.585 \\
DaSiamRPN~\cite{DaSiamRPN} &
  No &
  - &
  - &
  - &
  0.411 &
  0.610 &
  0.220 &
  0.326 &
  0.560 &
  0.340 \\
SiamRPN++~\cite{SiamRPN++} &
  No &
  0.733 &
  0.694 &
  0.800 &
  - &
  - &
  - &
  0.414 &
  0.600 &
  0.234 \\
ATOM~\cite{ATOM} &
  No &
  0.703 &
  0.648 &
  0.711 &
  - &
  - &
  - &
  0.401 &
  0.590 &
  0.204 \\
DiMP~\cite{DiMP} &
  No &
  0.740 &
  0.687 &
  0.801 &
  - &
  - &
  - &
  0.440 &
  0.597 &
  0.153 \\ \hline
KCF~\cite{kcf} &
  Yes &
  0.447 &
  0.419 &
  0.546 &
  0.192 &
  0.489 &
  0.569 &
  0.135 &
  0.447 &
  0.773 \\
ECO~\cite{eco} &
  Yes &
  0.561 &
  0.489 &
  0.621 &
  0.375 &
  0.550 &
  0.569 &
  0.280 &
  0.270 &
  0.480 \\
S2SiamFC~\cite{s2siamfc} &
  Yes &
  - &
  - &
  - &
  0.215 &
  0.493 &
  0.639 &
  0.180 &
  0.463 &
  0.782 \\
LUDT+~\cite{ludt} &
  Yes &
  0.563 &
  0.495 &
  0.633 &
  0.299 &
  0.570 &
  0.331 &
  0.230 &
  0.490 &
  0.412 \\
USOT~\cite{usot} &
  Yes &
  0.599 &
  0.551 &
  0.682 &
  0.351 &
  0.593 &
  0.336 &
  0.290 &
  0.564 &
  0.435 \\
USOT*~\cite{usot} &
  Yes &
  0.616 &
  0.566 &
  0.691 &
  {\color[HTML]{3166FF} 0.402} &
  {\color[HTML]{3166FF} 0.600} &
  0.233 &
  0.344 &
  {\color[HTML]{FE0000} 0.578} &
  {\color[HTML]{3166FF} 0.304} \\ \hline
ULAST*-off &
  Yes &
  {\color[HTML]{3166FF} 0.649} &
  {\color[HTML]{3166FF} 0.585} &
  {\color[HTML]{3166FF} 0.725} &
  0.397 &
  0.599 &
  {\color[HTML]{3166FF} 0.224} &
  {\color[HTML]{3166FF} 0.347} &
  0.569 &
  {\color[HTML]{3166FF} 0.304} \\
ULAST*-on &
  Yes &
  {\color[HTML]{FE0000} 0.654} &
  {\color[HTML]{FE0000} 0.592} &
  {\color[HTML]{FE0000} 0.732} &
  {\color[HTML]{FE0000} 0.417} &
  {\color[HTML]{FE0000} 0.603} &
  {\color[HTML]{FE0000} 0.214} &
  {\color[HTML]{FE0000} 0.355} &
  {\color[HTML]{3166FF} 0.571} &
  {\color[HTML]{FE0000} 0.286} \\ \hline
\end{tabular}
\vspace{-4mm}
\end{table*}

\noindent{\textbf{Network architecture.}}
The architecture of our network follows conventional Siamese trackers. We adopt the ResNet50 as the feature extractor. Features from layer 2, 3, 4 are used as inputs of the region proposal network (RPN), and these features are interpolated to the same spatial resolution. Also, a learned weight is used to aggregate these three correlation maps. The spatial resolution of search feature size is $31 \times 31$, the grid size in region mask operation is the same as this. 
The anchor scale of RPN is set as 8, and the anchor ratio is set as $ [0.33, 0.5, 1, 2, 3]$, ATSS~\cite{ATSS} is used to assign anchor label with the top-15 candidates. There are $K=3125$ candidate boxes in the prediction result of RPN.

\noindent{\textbf{Training details.}} Hyper-parameters for loss are set as $\lambda_1 = 10$, $\lambda_2=1.2$, $\lambda_c = 0.5$. In the region mask, $TH$ is set as $0$ to pass all predicted results. In the mask-guided re-weight strategy, we set hyper parameters as $\gamma=5$, $\alpha=7$ and $\beta=0.8s_{max}$, where $s_{max}$ denotes maximum score when generating the regional mask. The feature extractor of ResNet50 is initialized with Imagenet pretrain.
The legacy training of 5 epochs is adopted to initialize the tracker with the ability to locate the rough position of the target object. Then 20 epochs of cycle training with one template frame and three search frames is adopted to enable the tracker to track objects with strong variation in practical cases. In the training process, we set the batch size as 8, and an SGD optimizer is used. The initial learning rate is set as $1e^{-3}$, which is gradually decayed to $5e^{-5}$ in logspace.

\subsection{Comparison with SOTA}
We compare our proposed method with unsupervised and supervised methods on five challenging datasets, including OTB2015~\cite{OTB2015}, VOT2016~\cite{VOT2016}, VOT2018~\cite{VOT2018}, TrackingNet~\cite{TrackingNet} and LaSOT~\cite{LaSOT}. The offline tracking mode is denoted as ULAST*-off, while the online tracking mode with memory update is denoted as ULAST*-on.

\begin{table}[tbp]
\setlength{\abovecaptionskip}{0cm}
\centering
\caption{Evaluation results on OTB2015 and LaSOT datasets. }
\label{tab: otb_lasot_result}
\resizebox{1\linewidth}{!}{\begin{tabular}{cc|cccc}
\hline
                           &     & \multicolumn{2}{c}{OTB2015} & \multicolumn{2}{c}{LaSOT} \\
\multirow{-2}{*}{Tracker} &
  \multirow{-2}{*}{Unsup} &
  Suc$\uparrow$ &
  Pre$\uparrow$ &
  Suc$\uparrow$ &
  Pre$\uparrow$ \\ \hline
SiamFC~\cite{SiamFC}       & No  & 0.582        & 0.771        & 0.336       & 0.339       \\
SiamRPN~\cite{SiamRPN}     & No  & 0.637        & 0.851        & 0.411       & 0.380       \\
SiamRPN++~\cite{SiamRPN++} & No  & 0.696        & 0.923        & 0.495       & 0.493       \\ \hline
KCF~\cite{kcf}             & Yes & 0.485        & 0.696        & 0.178       & 0.166       \\
DSST~\cite{dsst}           & Yes & 0.518        & 0.689        & 0.207       & 0.189       \\
LUDT+~\cite{ludt}          & Yes & 0.639        & 0.843        & 0.305       & 0.288       \\
USOT~\cite{usot}           & Yes & 0.589        & 0.806        & 0.337       & 0.323       \\
USOT*~\cite{usot}          & Yes & 0.574        & 0.775        & 0.358       & 0.340       \\ \hline
ULAST*-off &
  Yes &
  {\color[HTML]{3166FF} 0.645} &
  {\color[HTML]{3166FF} 0.878} &
  {\color[HTML]{3166FF} 0.468} &
  {\color[HTML]{3166FF} 0.448} \\
ULAST*-on &
  Yes &
  {\color[HTML]{FE0000} 0.648} &
  {\color[HTML]{FE0000} 0.879} &
  {\color[HTML]{FE0000} 0.471} &
  {\color[HTML]{FE0000} 0.451} \\ \hline
\end{tabular}}
\vspace{-6mm}
\end{table}

\noindent\textbf{VOT2016.} There are totally 60 videos in this dataset. 
In this benchmark, three metrics are used to report tracking performance: Robustness (Rob), Accuracy (Acc) and Expected Average Overlap (EAO)~\cite{VOT2016}. 
Table~\ref{tab:all_performance} shows evaluation results of our tracker.
Without online update, the ULAST*-off achieves better robustness score than the USOT* with online update. 
When aided with online update, our ULAST*-on achieves the best tracking performance among these unsupervised trackers according to these three metrics. 

\noindent\textbf{VOT2018.} VOT2018 contains more challenging video sequences than VOT2016 datasets.
As shown in Table~\ref{tab:all_performance}, the ULAST*-off has an EAO score of $0.347$, which is superior to all unsupervised trackers without online update.
Furthermore, ULAST*-on can realize performance improvement with online update, achieving an EAO score of $0.355$.

%
\noindent\textbf{OTB2015.} OTB2015 contains 100 video sequences with various targets.
We compare our proposed method with eight representatives of supervised and unsupervised trackers, and evaluation results are shown in Table~\ref{tab: otb_lasot_result}. ULAST*-on achieves the best success and precision score among these unsupervised tracking with Success score of 0.648. And it is worth mentioning that our ULAST*-off outperforms methods~\cite{usot, ludt} that rely on online update.

\noindent\textbf{TrackingNet.} 
TrackingNet is a large-scale benchmark for tracking in the wild. In addition to the precision and success metric used in OTB2015, TrackingNet introduced another metric called normalized precision (NPre). The evaluation results are shown in Table~\ref{tab:all_performance}. ULAST*-off achieves a Success score of 0.649, outperforming the state-of-the-art unsupervised methods by a large margin.
And the ULAST*-on achieves a gain of 0.5 on success score when applying the online update with the CPT module.

\noindent\textbf{LaSOT.}
This benchmark dataset is the largest annotated one in the tracking community, consisting of 280 long videos sequences. 
The evaluation metric of this benchmark is the same as the TrackingNet, and the evaluation results are shown in Table~\ref{tab: otb_lasot_result}. ULAST*-off and ULAST*-on both achieve significant performance improvement on this benchmark with success score of 0.645 and 0.648.

\subsection{Ablation Study and Algorithm Analysis}
In this section, we conduct extensive experiments and give detailed analysis of our proposed method. For simplicity, region mask, region mask (detach head), Precise RoI-pooling, mask-guided loss re-weighting, residual connection are abbreviated as RM, RM(D), PrPool, ReLoss and Res respectively. Besides, models are all trained on LaSOT~\cite{LaSOT} and Got10k~\cite{got10k} with half iterations per epoch of the full version.

\begin{table}[h]
\vspace{-2mm}
\setlength{\abovecaptionskip}{0cm}
\centering
\caption{Quantitative analysis of CPT module and region mask with ULAST*-off on VOT2018 benchmark.}
\label{tab:comp_perf}
\resizebox{0.89\linewidth}{!}{\begin{tabular}{cccccccc}
\hline
RM & RM(D)      & PrPool     & CPT        & Res & Acc$\uparrow$                  & Rob$\downarrow$ & EAO $\uparrow$               \\ \hline
\checkmark &  &  & \checkmark &            & {\color[HTML]{3166FF} 0.560} & {\color[HTML]{FE0000} 0.272} & {\color[HTML]{FE0000} 0.346} \\
\checkmark &  &  & \checkmark & \checkmark & {\color[HTML]{3166FF} 0.560} & {\color[HTML]{3166FF} 0.304} & 0.317                        \\
   & \checkmark &            & \checkmark &     & 0.558                        & 0.337         & {\color[HTML]{3166FF} 0.322} \\
   &            & \checkmark & \checkmark &     & 0.552                        & 0.342         & 0.301                        \\
   &            & \checkmark &            &     & -                            & -             & -                            \\
   &            &            & \checkmark &     & {\color[HTML]{FE0000} 0.561} & 0.309         & 0.310                        \\ \hline
\end{tabular}}
\vspace{-2mm}
\end{table}

\noindent\textbf{Consistency Propagation Transformation(CPT).} For evaluating the contribution of the CPT module, we conduct three experiments.
Notably, the output of this module does not contain the common residual connection in attention-based operations. 
The motivation of this design is to exploit rich temporal feature of the target object, reducing the reliance on initial template feature.
To validate that, we carry out an experiment trained with residual connection in CPT module by adding the initial template feature to the output.
As the result shown in Table~\ref{tab:comp_perf}, the EAO score drops from $0.346$ to $0.317$ with this residual connection, demonstrating that involving the initial template will cause a performance drop.
Besides, we remove the region mask to evaluate the CPT module alone, the EAO score drops to 0.310, but compared to tracker trained with single frame (in Table~\ref{tab:reloss}, EAO=0.296), i.e. spatial self-supervision only, this model still has a performance improvement of 1.4 EAO. 
In addition, we try to replace the CPT module with PrPool. As mentioned above, the training pipeline is prone to breaks down (gradient explosion) in this setting for the incorrect prediction result of the tracker in intermediate frames.

\noindent\textbf{Region mask.} To better understand the proposed region mask, here we perform two experiments. 
First, the region mask operation is substituted by the Precise RoI-pooling~\cite{prpool} to select features from last frame. We pool the candidate features from top-3 scored predicted boxes after non-maximum suppression. Pooled features are then fused by averaging, as shown in Table~\ref{tab:comp_perf}, the EAO score drops to 0.301.
Second, we detach the gradient of candidate boxes on search region when generating the regional mask. In this form, the regional mask lost the ability to implicitly penalize the intermediate tracking errors. As the RM(D) option shown in Table~\ref{tab:comp_perf}, the EAO score drops from 0.346 to 0.322.

\label{subsec:ablation}
\begin{table}[tbp]
\setlength{\abovecaptionskip}{0cm}
\centering
\caption{Quantitative analysis of the mask-guided loss re-weighting on VOT2018 benchmark.}
\label{tab:reloss}
\resizebox{0.75\linewidth}{!}{\begin{tabular}{cccccc}
\hline
RM & CPT & ReLoss     & Acc$\uparrow$                & Rob$\downarrow$ & EAO $\uparrow$ \\ \hline
  &     &            & 0.559                        & 0.370           & 0.284          \\
  &     & \checkmark & {\color[HTML]{FE0000} 0.565} & 0.361           & 0.296          \\
\checkmark & \checkmark &            & 0.557                        & {\color[HTML]{3166FF} 0.309} & {\color[HTML]{3166FF} 0.331} \\
\checkmark & \checkmark & \checkmark & {\color[HTML]{3166FF} 0.560} & {\color[HTML]{FE0000} 0.272} & {\color[HTML]{FE0000} 0.346} \\ \hline
\end{tabular}}
\vspace{-2mm}
\end{table}

\begin{table}[tbp]
\setlength{\abovecaptionskip}{0cm}
\centering
\caption{Quantitative analysis of pretrain feature impact on OTB2015 and LaSOT benchmark. * denotes the ULAST model trained with ImageNet pretrain.}
\label{tab:pretrain_analysis}
\resizebox{0.7\linewidth}{!}{\begin{tabular}{ccccc}
\hline
                               & \multicolumn{2}{c}{OTB2015}                  & \multicolumn{2}{c}{LaSOT}                    \\
\multirow{-2}{*}{setting}      & Suc$\uparrow$ & Pre$\uparrow$                & Suc$\uparrow$                & Pre$\uparrow$ \\ \hline
\multicolumn{1}{c|}{ULAST-on}  & 0.610         & 0.811                        & {\color[HTML]{000000} 0.433} & 0.407         \\
\multicolumn{1}{c|}{ULAST-off} & 0.607         & {\color[HTML]{3166FF} 0.812} & 0.429                        & 0.405         \\
\multicolumn{1}{c|}{ULAST$^*$-on}  & {\color[HTML]{FE0000} 0.643} & {\color[HTML]{FE0000} 0.862} & {\color[HTML]{FE0000} 0.442} & {\color[HTML]{FE0000} 0.418} \\
\multicolumn{1}{c|}{ULAST$^*$-off} & {\color[HTML]{3166FF} 0.639} & {\color[HTML]{FE0000} 0.862} & {\color[HTML]{3166FF} 0.436} & {\color[HTML]{3166FF} 0.410} \\ \hline
\end{tabular}}
\vspace{-6mm}
\end{table}

\noindent\textbf{Mask guided sample re-weighting.}
Here we conduct experiments on cycle training and legacy training (training by template-search pairs from single frames) to validate the effectiveness of this strategy. As shown in Table~\ref{tab:reloss}, with this strategy, both of the performances are boosted in EAO: from 0.331 to 0.346 and from 0.284 to 0.296 for cycle training and legacy only training, respectively.

\noindent\textbf{Impact of ImageNet pretrain.} 
For fair comparison to existing unsupervised tracking works, here we train our tracker from scratch. 
Detailly, we substitute ResNet50~\cite{RES50} pretrained on ImageNet classification as DenseCL~\cite{denscl} pretrain, which is self-supervised trained with contrastive learning.
Then the trained model is evaluated on OTB2015 and LaSOT benchmark datasets for comparison, results are shown in Table~\ref{tab:pretrain_analysis}.
The performance of the ULAST-on and ULAST-off have a relatively slight performance drop compared to ULAST* model.
It suggests that a better representation can contribute to unsupervised visual tracking.

\section{Conclusion}

In this paper, we propose a novel framework for unsupervised tracking called ULAST.
To generate reliable template kernel in the forward propagation process and thus enable our framework to be trained with cycle consistency, we first propose a consistency transformation.
In the backward propagation process, we propose a region mask operation to implicitly penalize tracking error on intermediate frames.
Besides, a mask-guided loss re-weighting strategy is proposed to assign dynamic weight to the loss from samples of various pseudo label qualities. 
With these proposed components, our ULAST can fully explore temporal supervision in unsupervised tracking process and achieves state-of-the-art performance.

\textbf{Acknowledgement}: Wanli Ouyang was supported by the Australian Research Council Grant DP200103223, Australian Medical Research Future Fund MRFAI000085, and CRC-P Smart Material Recovery Facility (SMRF) – Curby Soft Plastics.
\clearpage

\section{Discussion and limitations}
Despite proposed framework is effective to learn a better tracker from temporal self-supervision, the pipeline still relies on pseudo labels in initial frames generated by unsupervised optical flow model.
As the relationship between a better initialization method and better tracker is a chicken-egg conundrum in this formulation, it's still a remaining problem about how to chain the initialization methods and unsupervised tracking into an end-to-end trainable pipeline.

\begin{table}[bp]
\vspace{-4mm}
\centering
\caption{Quantitative analysis on VOT2018 benchmark}
\resizebox{0.7\linewidth}{!}{\begin{tabular}{cccc}
\hline
template boxes              & Acc$\uparrow$ & Rob$\downarrow$ & EAO$\uparrow$ \\ \hline
\textit{ground-truth} & 0.600       & 0.234         & 0.414         \\
\textit{jittered gt}  & 0.565       & 0.355         & 0.298         \\ \hline
\end{tabular}}
\label{tab:q_analysis}
\end{table}

\section{Quantitative analysis of misalignments}
Here we conduct experiments to give a quantitative analysis on the misalignments in conventional cycle training of unsupervised visual tracking.
In our manuscript, we claim that the misalignment in cycle training severely hinders the performance of unsupervised visual tracking. 
Detailly, the source of this misalignment is the mismatch between the internal template and search region feature. For delving deep into this insight, we conduct a quantitative analysis of the impact of this misalignment.
We choose the classical Siamese tracker, SiamRPN++~\cite{SiamRPN++}, as the baseline. For simplification, the mismatch in intermediate frames, which may be produced by forward tracking errors or initialization bias, is simulated as noises added to ground-truth for cropping the template patches in SiamRPN++~\cite{SiamRPN++}. Specifically, we add noises to template patches with the following operation. 
Let us denote the ground-truth boxes in template frames as $(cx, cy, w, h)$, where $(cx, cy)$ in the center coordinates of bounding boxes, $w$ and $h$ are the width and height of the boxes respectively.
The jittered template bounding boxes can be denoted as $(cx + \sigma_1 w, cy + \sigma_2 h, (1 + \sigma_3) w, (1 + \sigma_4) h)$, where $\sigma_{k}$ denotes random number between $-0.5$ and $0.5$ generated from uniform distribution.

In training phase, the model trained with noisy template boxes is hard to convergent on the regression branch. We evaluate the tracking performance on VOT2018~\cite{VOT2018} benchmark dataset, as shown in Table~\ref{tab:q_analysis}, when template boxes are jittered, the performance of the tracker will drop with a substantial gap in the EAO metric.

\section{More discussion about proposed component}
\subsection{Threshold in region mask}
\begin{figure*}[p]
     \centering
     \begin{subfigure}[b]{0.48\textwidth}
         \centering
         \includegraphics[width=\textwidth]{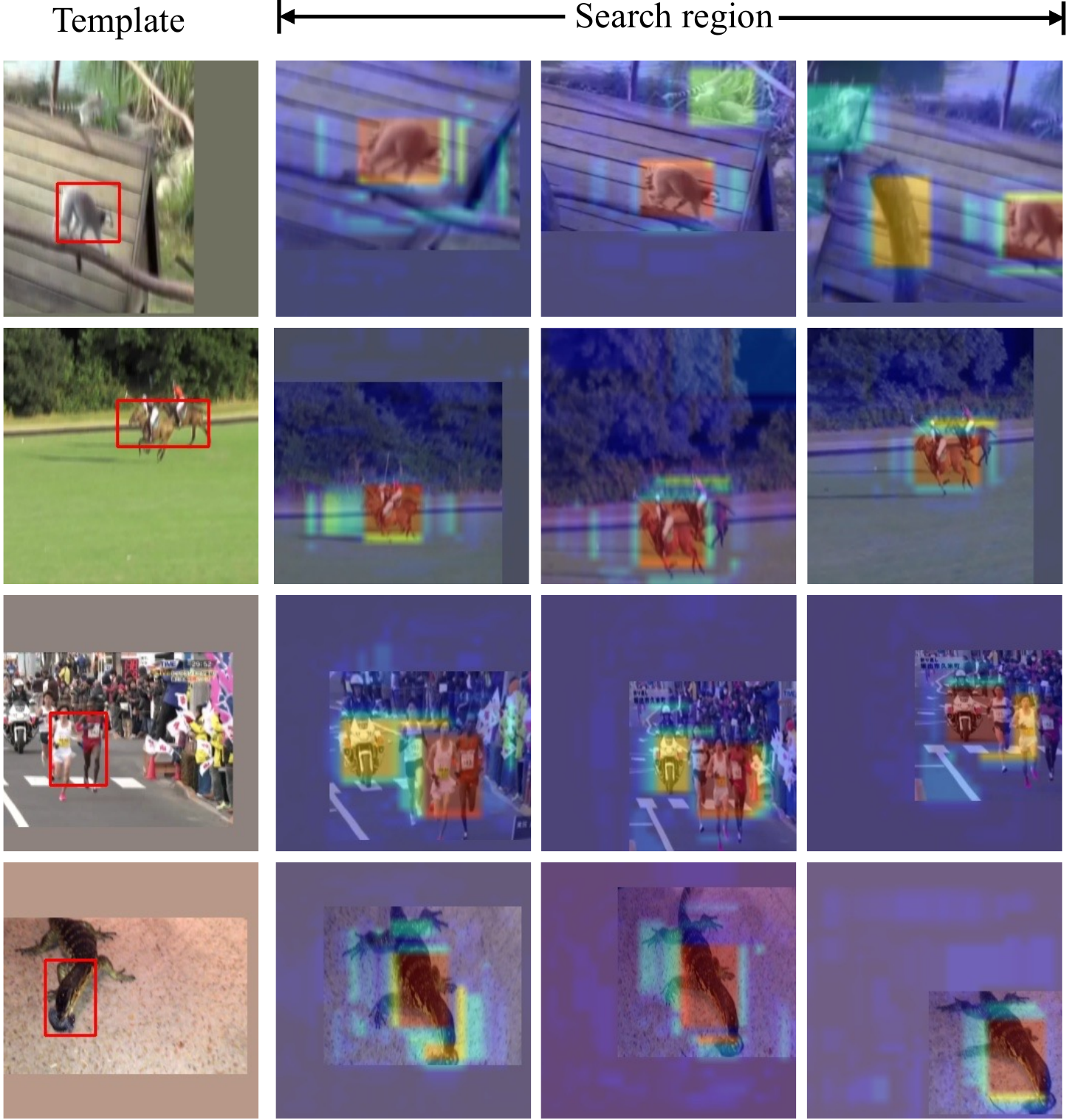}
         \caption{$TH=0.0$}
     \end{subfigure}
     \hfill
     \begin{subfigure}[b]{0.48\textwidth}
         \centering
         \includegraphics[width=\textwidth]{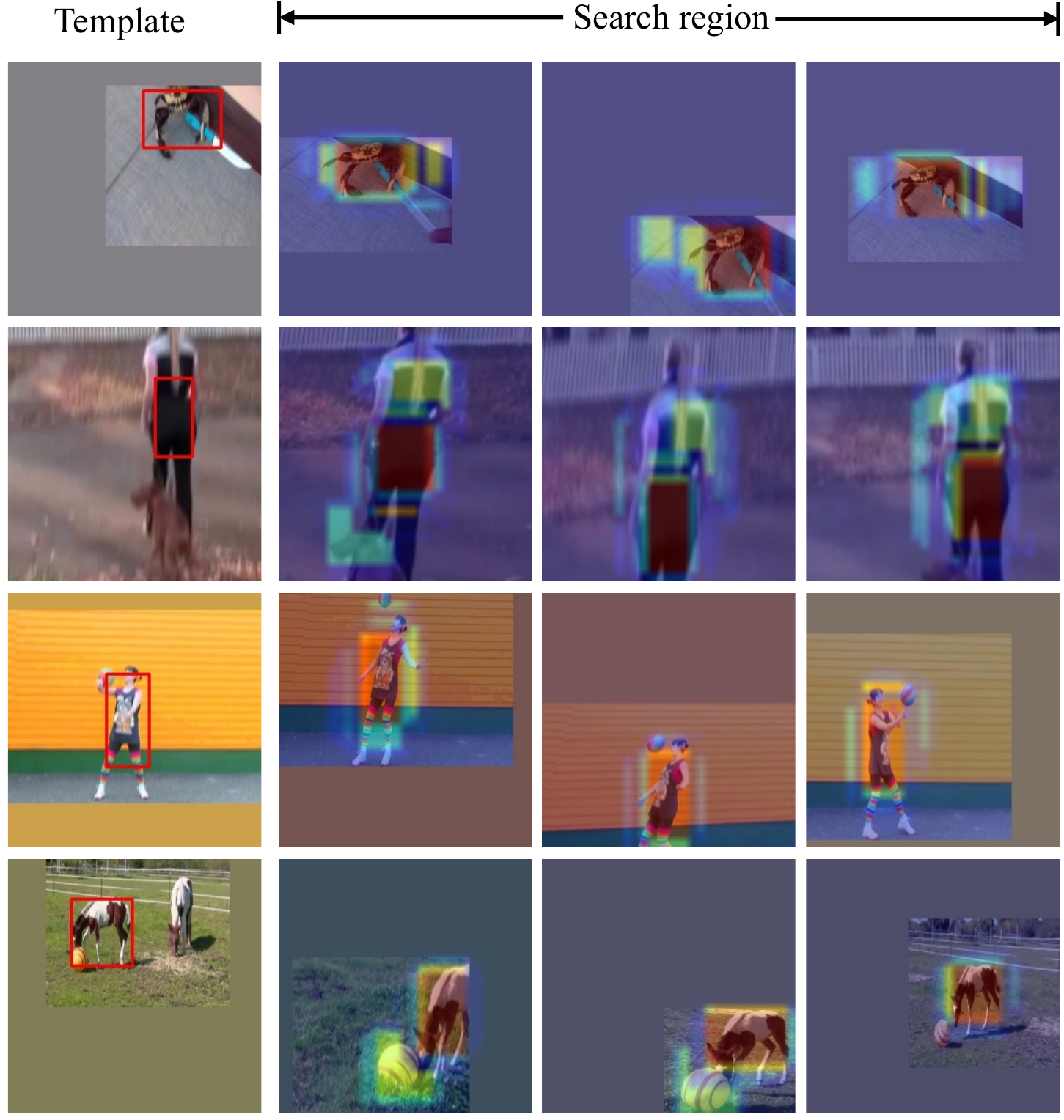}
         \caption{$TH=0.4$}
     \end{subfigure}
     \hfill
     \begin{subfigure}[b]{0.48\textwidth}
         \centering
         \includegraphics[width=\textwidth]{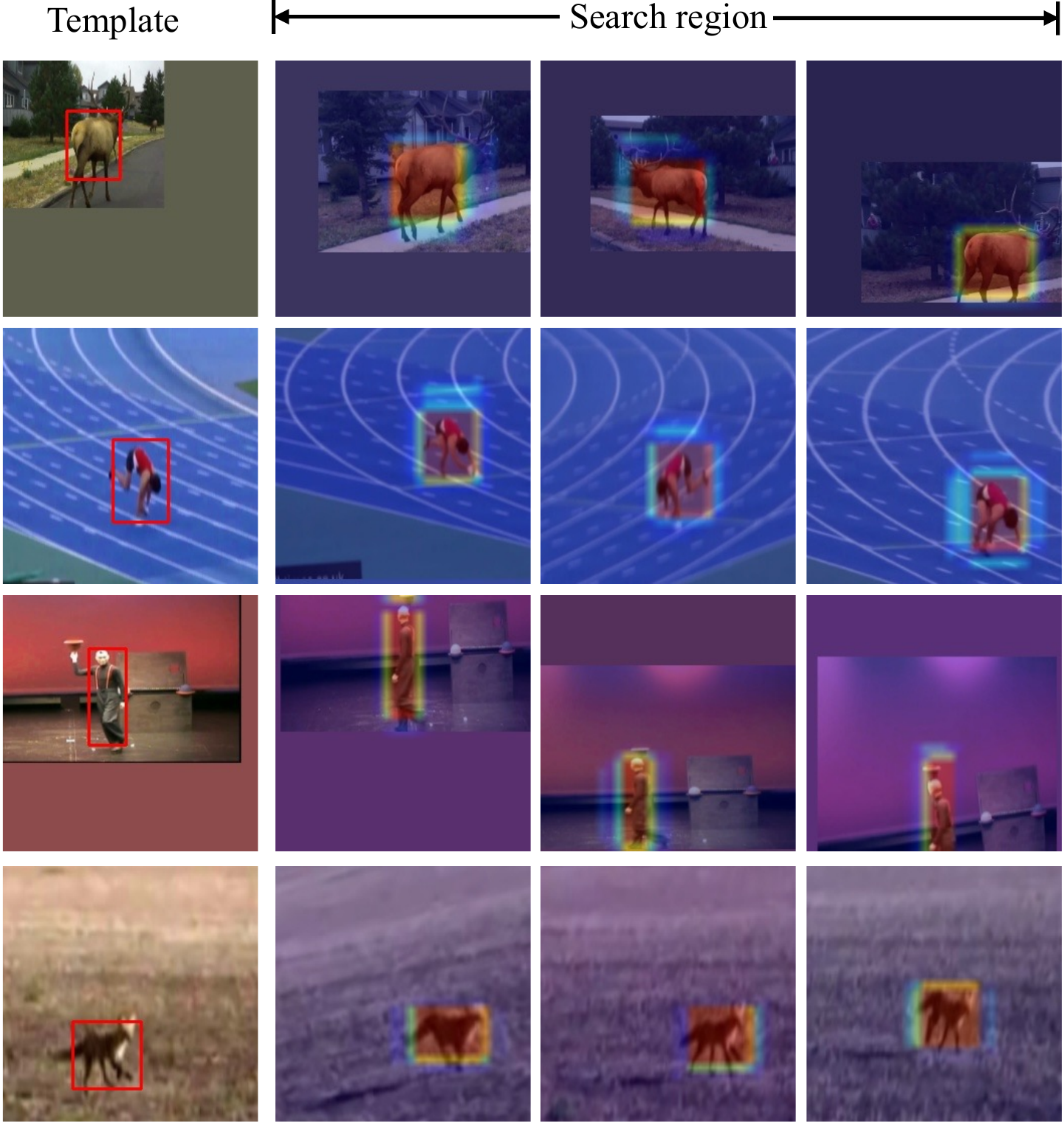}
         \caption{$TH=0.6$}
     \end{subfigure}
     \hfill
    \begin{subfigure}[b]{0.48\textwidth}
        \centering
         \includegraphics[width=\textwidth]{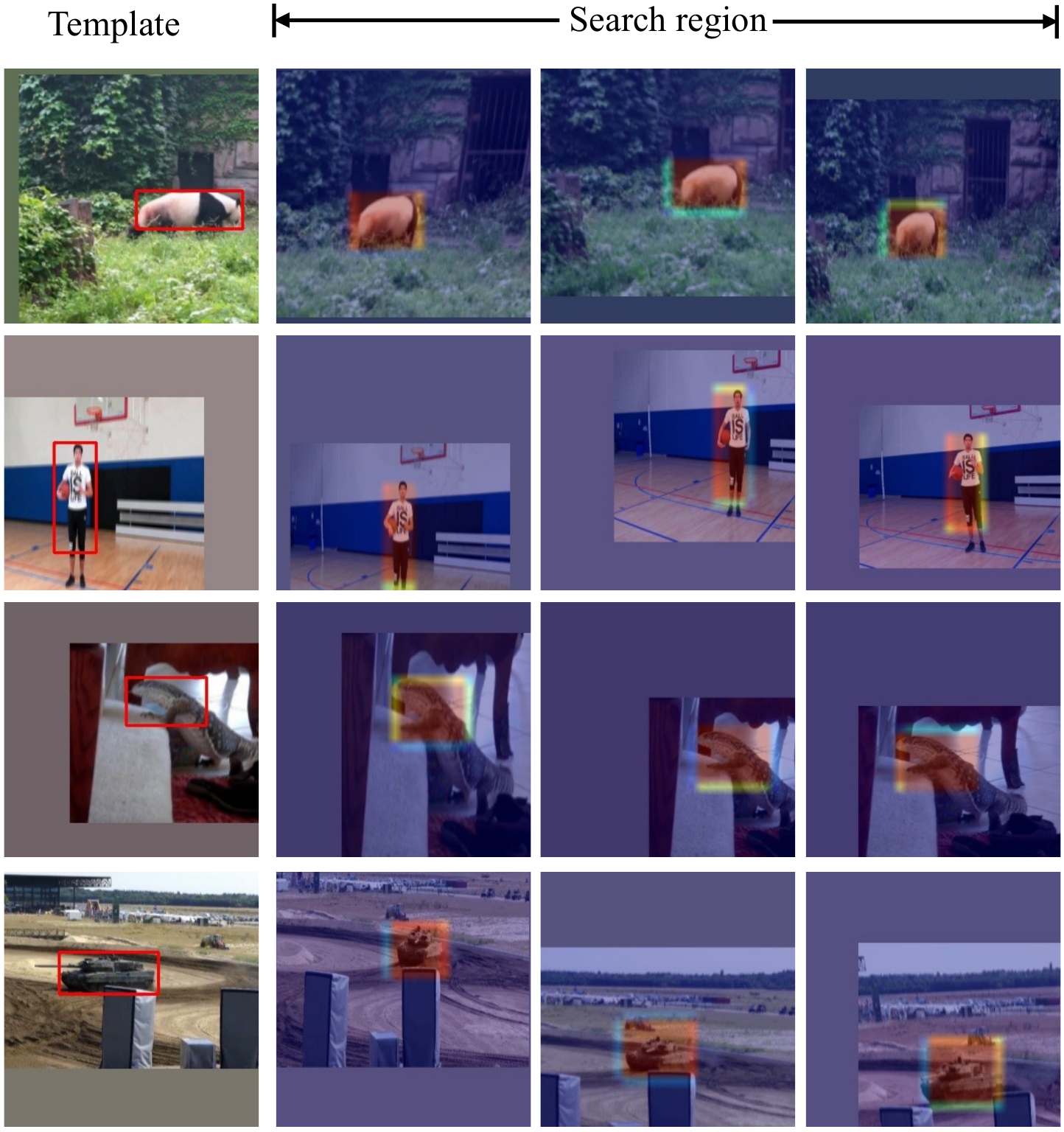}
         \caption{$TH=0.9$}
    \end{subfigure}
    \hfill
        \caption{Visualization of regional mask on training samples with different positive threshold value}
        \label{fig:mask_TH}
\end{figure*}

Our proposed region mask is a customized operation for unsupervised visual tracking in cycle training, which penalizes tracking errors on intermediate frames by making the coordinates differentiable. 
As claimed in our manuscripts, when compared to conventional feature selection operation like PrPool~\cite{prpool}, this operation is efficient to select features from last search region feature based on the output of region proposal network (RPN), denoted as $P_{cls}$ and $P_{reg}$. 
Specifically, we set boxes number as 3125 ($25 \times 25 \times 5$), the same as the total number of predicted boxes of RPN. And the positive threshold (denoted as $TH$ in the manuscript) is set as 0 for passing all predicted boxes.
Besides, we also visualize the regional mask propagation in training samples with different positive thresholds.
The regional masks on search region images with four different threshold values are shown in Fig \ref{fig:mask_TH}.
When the value of this threshold increases, the region mask tend to filter out more predicted boxes with low confidence scores, which results in less information propagating between frames.
Based on this observation, we always set $TH=0$ in training phase, for propagating more information between frames. In addition, when proposal boxes with top confidence scores are wrong, region mask with a lower positive threshold is more likely to select the correct proposal box.

For better understanding, we give a quantitative analysis of this threshold $TH$ in training phase.
As ablation studies with other thresholds shown in Table~\ref{tab:rm_thres}, the EAO scores drop significantly on the EAO scores as threshold increases.
Every region map is multiplied with its confidence score when generating the region mask, thus the samples with low confidence essentially have smaller gradient.
However, considering all boxes (including a large amount of non-target regions) essentially accumulates abundant training samples.
As another aspect, such noisy region masks in training enforce the tracker to learn better discriminating abilities, i.e., predict higher confidence scores on the foreground (target) area and lower scores on the background area.
While in online tracking phase, we cache search regions features of high confidence and corresponding regional mask for updating in a memory queue. When retrieving the memory kernel with the CPT module, we set a higher threshold to filter similar distractors features.
\begin{table}[tbp]
\vspace{-6mm}
\centering
\caption{Ablation study of the threshold of region mask}
\label{tab:rm_thres}
\resizebox{0.6\linewidth}{!}{\begin{tabular}{cccc}
\hline
Threshold & ACC $\uparrow$               & Rob $\downarrow$             & EAO $\uparrow$               \\ \hline
0.0       & {\color[HTML]{FE0000} 0.560} & {\color[HTML]{FE0000} 0.272} & {\color[HTML]{FE0000} 0.346} \\
0.5       & {\color[HTML]{000000} 0.550} & {\color[HTML]{3531FF} 0.323} & {\color[HTML]{3531FF} 0.327} \\
0.9       & {\color[HTML]{3531FF} 0.559} & 0.342                        & 0.303                        \\ \hline
\end{tabular}}
\vspace{-2mm}
\end{table}

\begin{table}[t]
\centering
\caption{Ablation study of the CPT module on VOT2018}
\label{tab:cpt_ablation}
\resizebox{0.6\linewidth}{!}{\begin{tabular}{cccc}
\hline
\multicolumn{1}{l}{Settings} & \multicolumn{1}{l}{ACC $\uparrow$} & \multicolumn{1}{l}{Rob $\downarrow$} & \multicolumn{1}{l}{EAO $\uparrow$} \\ \hline
\textit{LT + ST} & {\color[HTML]{3531FF} 0.560} & {\color[HTML]{FE0000} 0.272} & {\color[HTML]{FE0000} 0.346} \\
\textit{LT}      & {\color[HTML]{FE0000} 0.562} & {\color[HTML]{3531FF} 0.318} & {\color[HTML]{3531FF} 0.330} \\
\textit{ST}      & 0.557                        & 0.328                        & {\color[HTML]{000000} 0.324} \\ \hline
\end{tabular}}
\end{table}

\begin{figure}[t]
\centering
\includegraphics[width=0.90\linewidth]{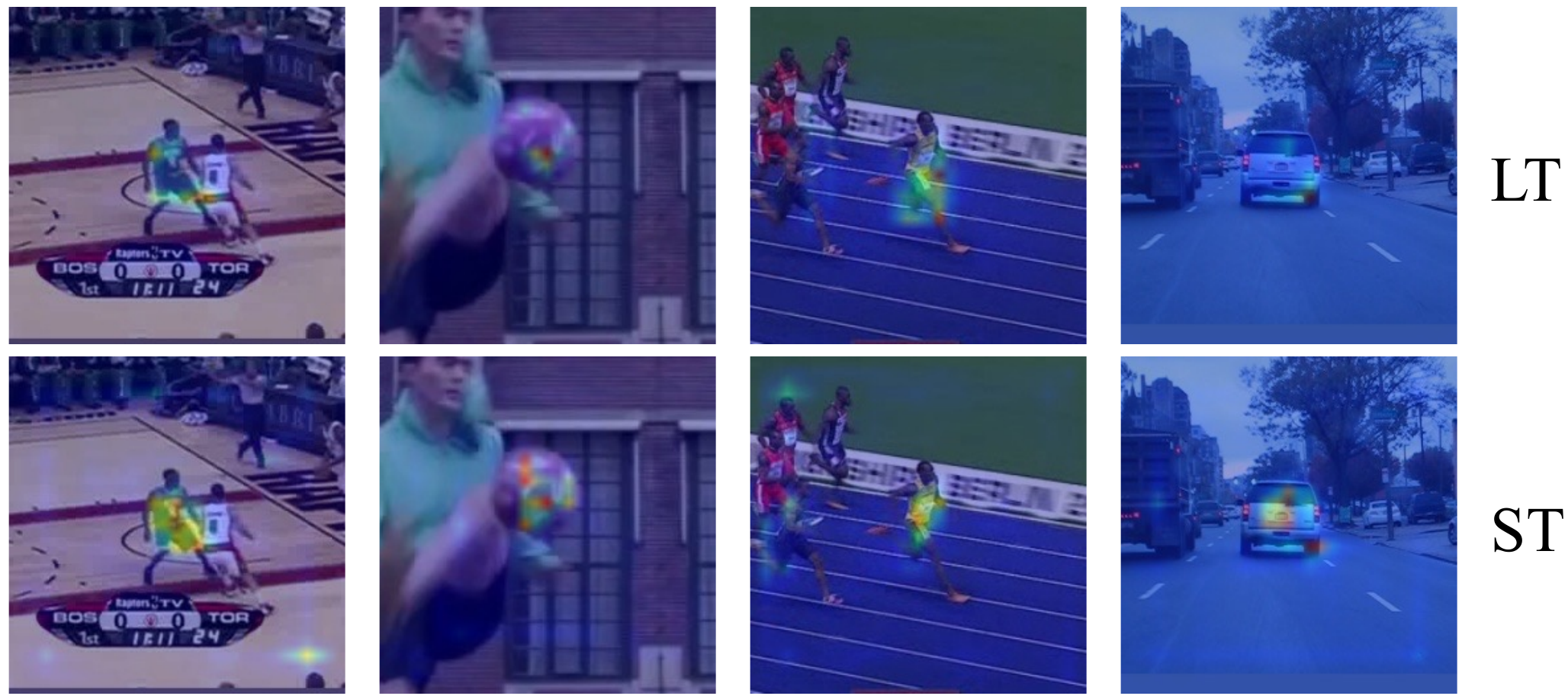}
\caption{Visualization of attention maps for LT and ST queries.}
\label{fig:attn_map}
\vspace{-6mm}
\end{figure}

\subsection{Long/short term in CPT module}
For training with cycle consistency, it is required to track videos frames as a cycle.
Previous works generated template kernels by RoI-Pooling on the top-1 proposal in search frames. 
If this top-1 proposal is wrong, especially in initial stage, then the generated template kernel becomes too noisy for the tracker to track back.
With proposed CPT module, multiple possible matched regions can be used for generating reliable template features between frames.
Here we visualize attention maps on search region for long and short term queries in Fig~\ref{fig:attn_map}, long-term (LT) queries have higher responses on invariant areas of the target, while short-term (ST) queries have higher responses on variant target areas as they are intended for retrieving the most recent target features.
Besides, we present ablation study on the long/short term query of CPT module that shown in Table~\ref{tab:cpt_ablation}. It shows that the combination of long-term and short-term queries performs better on EAO scores than using single term. 

\clearpage
{\small
\bibliographystyle{ieee_fullname}
\bibliography{ulast}
}

\end{document}